\documentclass[letterpaper, 10 pt, conference]{ieeeconf} 

\IEEEoverridecommandlockouts     
\usepackage{cite}
\usepackage{amsmath,amssymb,amsfonts}
\usepackage{mathtools}
\usepackage{graphicx}
\usepackage{textcomp}
\usepackage{xcolor}
\usepackage{subfigure}
\usepackage{placeins}
\usepackage{float}
\usepackage{hyperref} 
\usepackage{bm}
\usepackage{optidef}

\usepackage[font={small}]{caption}

\usepackage[noend]{algorithmic}
\usepackage[ruled,noend]{algorithm2e}
\SetAlCapHSkip{0pt}

\newcommand{\ie}{i.e.,\ }

\DeclarePairedDelimiter{\norm}{\lVert}{\rVert}

\title{\LARGE \bf Multi-Robot Localization and Target Tracking with \\ Connectivity Maintenance and Collision Avoidance}

\author{Rahul Zahroof*$^{1}$, Jiazhen Liu*$^{1}$, Lifeng Zhou$^{2}$, Vijay Kumar$^{1}$
\thanks{*Equally contributed.}
\thanks{$^{1}$The authors are are with the GRASP Laboratory, University of Pennsylvania, Philadelphia, PA 19104, USA (email: {\tt\small\{rahulz, jzliu, kumar\}@seas.upenn.edu}).}
\thanks{$^{2}$The author is with the Department of Electrical and Computer Engineering, Drexel University, Philadelphia, PA 19104, USA (email: {\tt\small lz457@drexel.edu}).}
\thanks{This research was sponsored by the Army Research Lab through ARL DCIST CRA W911NF-17-2-0181.}}

\begin{document}

\maketitle

\begin{abstract}
We study the problem that requires a team of robots to perform joint localization and target tracking task while ensuring team connectivity and collision avoidance. The problem can be formalized as a nonlinear, non-convex optimization program, which is typically hard to solve. To this end, we design a two-staged approach that utilizes a greedy algorithm to optimize the joint localization and target tracking performance and applies control barrier functions to ensure safety constraints, \ie maintaining connectivity of the robot team and preventing inter-robot collisions. Simulated Gazebo experiments verify the effectiveness of the proposed approach. We further compare our greedy algorithm to a non-linear optimization solver and a random algorithm, in terms of the joint localization and tracking quality as well as the computation time. The results demonstrate that our greedy algorithm achieves high task quality and runs efficiently. 
\end{abstract}



\section{Introduction}
\label{sec:intro}
Multi-robot systems are attracting increasing research attention due to their wide applications in fields such as search and rescue\cite{9220149}, environment monitoring~\cite{5597912}, exploration~\cite{1677947}, and many more. In most applications, the multi-robot team is equipped with a suite of sensors to perform team-level tasks. To optimize the task performance, the robots need to actively reconfigure their positions as well as coordinate with each other. The specific task motivating this paper is multi-robot multi-target tracking, or using multiple robots to track the positions of multiple targets. In contrast to the sole problem of target tracking, which generally assumes the true positions of robots to be known a priori~\cite{zhou2011multirobot, bar2004estimation,8466114,zhou2019sensor}, our setting requires estimating both the robots' and targets' positions using sensors mounted on the robots. This joint task of localization and target tracking is further complicated by the fact that the robots often have a limited communication range. If a certain robot is not within the communication range of any of its teammates, localization and tracking performance would deteriorate since the overall estimation accuracy heavily depends on the robots exchanging information with each other. For instance, a robot out of contact with other teammates may suffer from poor localization. Due to the lack of knowledge about its accurate position, the robot's target tracking performance may also degrade. Therefore, the communication network formed by the robot team should always remain connected. Meanwhile, inter-robot collision avoidance should be avoided by ensuring that the robots maintain a minimum safety distance between each other.


Our major contributions are the formulation of the joint problem of localization and target tracking as a nonlinear, nonconvex optimization program and the development of a two-staged approach for solving the program. In the first stage, we design a greedy algorithm that optimizes the performance of joint localization and multi-target tracking without considering any constraints. Then in the second stage, we leverage control barrier functions (CBFs) to ensure safety constraints such as connectivity maintenance and collision avoidance. The proposed greedy algorithm achieves high performance on the joint task. Furthermore, it runs in polynomial time and thus favorably scales up to larger team sizes. The upcoming sections are arranged as follows. Section~\ref{sec:related_work} grounds our work on the foundation of previous literature. Section~\ref{sec:problem-formulate} details notation conventions and formal definitions for the joint self-localization and target tracking problem. The main greedy algorithm, along with our methods for maintaining team connectivity and computing estimates, is described in Sec.~\ref{sec:approach}. We present both qualitative illustrations and quantitative comparisons in Sec.~\ref{sec:results}. Finally, Sec.~\ref{sec:conclusion} concludes the paper and proposes several future extensions.

\section{Related Work}
\label{sec:related_work}
The joint task of self-localization and target tracking has been previously addressed by a sizable amount of literature~\cite{8864098, moosmann2013joint, kantas2012distributed, 6489214}. Concretely,~\cite{8864098} focuses on the case where the association between target measurements and target identities is unknown. A novel decentralized method is proposed to deal with an unknown and time-varying number of targets under association uncertainty. The generalized approach proposed in~\cite{moosmann2013joint} for joint self-localization and tracking of generic 3D objects is applicable to any type of environment. In~\cite{kantas2012distributed}, the self-localization problem is cast as a static parameter estimation problem for Hidden Markov Models. Decentralized adaptations of the Recursive Maximum Likelihood and online Expectation-Maximization algorithms are used to address self-localization along with target tracking. These works primarily focus on algorithm design for either localization or tracking, leaving out safety guarantees such as network connectivity maintenance, which is an essential component if a multi-robot team is to be deployed for executing practical tasks. Moreover, considering that self-localization and target tracking could both be seen as iterative state estimation problems, various filtering algorithms have been applied to deal with uncertainty in the estimates~\cite{tallamraju2018decentralized, wakulicz2021active, dias2013cooperative}. Localization and target tracking have also been explored in more challenging scenarios with limited GPS~\cite{5174796, 6858831, 6719359}. 

Besides pursuing accurate localization and target tracking, safety-critical constraints such as connectivity maintenance and collision avoidance should be properly considered. To this end,~\cite{ames2019control} proposes a CBF for connectivity maintenance. It provides an elaborate discussion on the CBF and its relationship with Lyapunov control. In~\cite{capelli2021decentralized}, the connected region is formulated as a safety set, and the CBF is utilized to render the set forward invariant. This guarantees the connectivity of the network throughout the duration of the task as long as the robots are initially connected. The work~\cite{sabattini2013distributed} approaches connectivity maintenance by considering spectral graph properties such as algebraic connectivity. In~\cite{ji2007distributed}, control actions are designed according to the weights assigned to graph edges in order to ensure connectivity.

\section{Problem Formulation}
\label{sec:problem-formulate}

We consider that a team of robots is tasked to track multiple dynamic targets. We assume the prior position estimates of all robots and targets to be known. However, such prior knowledge is noisy and not accurate. The goal of the robots is to actively reconfigure their positions to minimize the uncertainties originating from target tracking and localization, while maintaining team connectivity and avoiding collisions. In this section, we first describe the notations to be used throughout the paper, then introduce the modeling of robots and targets, and finally define the joint task of target tracking and localization as an optimization problem. 



\subsection{Notation}
Capital letters in boldface are used to denote matrices; lower-case letters in boldface represent vectors; lower-case letters of regular font are scalars. Subscripts are used to indicate the relevant group (the set of targets denoted by $\mathcal{T}$, robots denoted by $\mathcal{R}$), measurement type, and vector/matrix indices. Superscripts indicate the time step at which a variable is computed. The overhead bar indicates a prior estimate, and the overhead hat indicates a posterior estimate. Vectors and matrices described as sequences of vectors/matrices in the form $\mathbf{M} = \left[\mathbf{M}_1, \mathbf{M}_2,\cdots ,\mathbf{M}_N\right]$ imply horizontal concatenation. The matrix format $\mathbf{M} = \text{blockdiag}\left(\mathbf{M}_1, \mathbf{M}_2, \cdots, \mathbf{M}_N\right)$ describes a block-diagonal construction of matrix $\mathbf{M}$ with the listed matrices placed along its diagonal. $\|\cdot\|_2$ represents 2-norm of vectors. 

\subsection{Modeling of Robots and Targets}
\paragraph{Robots} We consider a team of $N$ robots, indexed by $\mathcal{R} = \{1, \cdots, N\}$. Each robot $i \in \mathcal{R}$ has the following discrete motion model:
\begin{equation}
    \bm{x}^{t}_{R_i} = \mathbf{A}_{R_i}\bm{x}_{R_i}^{t-1} + \mathbf{B}_{R_i}\bm{u}_{R_i}^{t-1} + \mathbf{G}_{R_i}\bm{w}_{d,R_i}^{t-1}, 
    \label{eq:robot_model}
\end{equation}
where $t$ denotes the current time step and $\mathbf{A}_{R_i}$ and $\mathbf{B}_{R_i}$ are the process and control matrices, respectively. The term $\bm{u}_{R_i}$ denotes the control input for robot $i$. $\mathbf{G}_{R_i}$ is the gain matrix to amplify noise for robot $i$, and $\bm{w}_{d}$ is a white Gaussian noise with zero mean and covariance $\mathbf{Q}_{d}^{t} = \mathbb{E}[\bm{w}^{t}_{d}\left(\bm{w}^{t}_{d}\right)^T]$.

\paragraph{Targets} We consider $M$ targets, indexed by $\mathcal{T} = \{1, \cdots, M\}$. Each target $i \in \mathcal{T}$ has the following discrete motion model:
\begin{equation}
    \bm{x}^{t}_{T_i} = \mathbf{A}_{T_i}\bm{x}_{T_i}^{t-1} + \mathbf{B}_{T_i}\bm{u}_{T_i}^{t-1} + \mathbf{G}_{T_i}\bm{w}_{d,T_i}^{t-1}, 
    \label{eq:target_model}
\end{equation}
where the control input $\bm{u}_{T_i}$ is predefined.
\paragraph{Sensing} We consider that each robot $i \in \mathcal{R}$ makes observations of each target $j \in \mathcal{T}$ according to the following measurement model:
\begin{equation}
    \bm{z}_{R_i, T_j}^{t} = \mathbf{h}\left(\bm{x}_{R_i}^t, \bm{x}_{T_j}^t\right) + \bm{n}_{R_i, T_j}^t,
    \label{eq:target_measurement_model}
\end{equation}
where $\mathbf{h}$ is the measurement function (e.g., range and bearing measurement) and $\bm{n}_{R_i, T_j}^t$ is a zero-mean white Gaussian noise term with covariance $\mathbf{R}_{R_i, T_j}$. The measurement function $\mathbf{h}$, the measurement noise $\bm{n}_{R_i, T_j}^t$, and the noise covariance $\mathbf{R}_{R_i, T_j}$ all depend on the states of the robots and targets.  Each robot $i \in \mathcal{R}$ also makes observations of itself and every other robot $j \in \mathcal{R}, j\neq i$ using the same model:

\begin{equation}
    \bm{z}_{R_i, R_j}^{t} = \mathbf{h}\left(\bm{x}_{R_i}^t, \bm{x}_{R_j}^t\right) + \bm{n}_{R_i, R_j}^t,
    \label{eq:robot_measurement_model}
\end{equation}
where $\bm{n}_{R_i, R_j}^t$ is a zero-mean white Gaussian
noise term with covariance $\mathbf{R}_{R_i, R_j}$. Self-observation relying on GPS is denoted by $\bm{z}_{R_i, R_i}^{t}$ with a zero-mean white Gaussian
noise term $\bm{n}_{R_i, R_i}^{t}$ which has covariance $\mathbf{R}_{R_i, R_i}$.
\paragraph{Communication} Each robot communicates only with the robots within a prescribed communication range $r_c$. We assume all robots to have the same communication range $r_c$. Based on this range, an (undirected) communication graph $G^t=\{\mathcal{R},\mathcal{E}^t\}$ is induced at step $t$, with the robots $\mathcal{R}$ as nodes. Edges $\mathcal{E}^t$ between nodes are defined such that $(i,j)\in\mathcal{E}^t$ if and only if $\|\bm{x}^{t}_{R_i}-\bm{x}^{t}_{R_j}\|_2\leq r_c$. We use $\mathbf{L}^t$ and $\bm{\lambda}_2^{t}$ to denote the weighted Laplacian matrix of $G^t$ and the second smallest eigenvalue of $\mathbf{L}^t$, respectively. 


\subsection{Problem Definition}
The problem of optimizing the quality of joint target tracking and localization while ensuring team connectivity and collision avoidance can be formally defined as the following optimization program: 
\begin{mini}|s|
{\bm{u}^{t-1}_\mathcal{R}}{\text{Tr}(\hat{\mathbf{P}}^t)}
{}{}
\addConstraint{ \bm{\lambda}_2^{t} > 0}
\addConstraint{\|\mathbf{x}_{R_i}^t - \mathbf{x}^{t}_{R_j}\|_2 \geq d_{\min}, ~\forall i,j \in \mathcal{R} } 
\addConstraint{\|\bm{u}_{R_i}^{t-1}\|_2 < u_{\max},  ~\forall i \in \mathcal{R}.} 
\label{opt1}
\end{mini}

The objective is to minimize the uncertainty in the positions of both the robots and targets. We quantify the uncertainty as the trace of the joint posterior covariance matrix of the robots and targets, denoted by $\hat{\mathbf{P}}^t = \text{blockdiag}(\hat{\mathbf{P}}^t_\mathcal{R}, \hat{\mathbf{P}}^t_\mathcal{T})$ with $\hat{\mathbf{P}}^t_\mathcal{R}$ and $\hat{\mathbf{P}}^t_\mathcal{T}$ being the covariance matrices of the robots and targets, respectively. In addition, the program includes three constraints. The first constraint ensures the connectivity of robots by making the second smallest eigenvalue larger than zero (refer to~\cite{godsil2001algebraic} for further details). The second constraint guarantees that the distance between any two robots is larger than the desired threshold $d$, thus ensuring inter-robot collision avoidance. The third constraint specifies the maximum motion ability of each robot. The optimal solution of this program (i.e., the robots' optimal control inputs) drives the robot team to a configuration where the uncertainty \textit{at the new} configuration is minimized. 

Notably, both the objective (i.e., trace) and constraints (i.e., connectivity and collision avoidance) are nonlinear and nonconvex in the decision variable (i.e., control inputs), which makes the program hard to solve. To tackle Program~\ref{opt1}, we present a two-staged method, with the details presented in the next section.  

\section{Approach}
\label{sec:approach}
As introduced in Sec.~\ref{sec:problem-formulate}, the joint task of localization and target tracking is in general a nonlinear and nonconvex optimization program and thus is challenging to solve. To this end, we present a two-stage approach. In the first stage, we focus on optimizing the objective (i.e., minimizing the estimated trace of covariance) without considering the three constraints. We design a greedy algorithm (Alg.~\ref{algo1}) to solve the unconstrained optimization problem to generate desired control inputs for the robots. In the second stage, we utilize CBFs to generate robots' actual control inputs that are close to their desired control inputs while ensuring the three constraints. After executing the actual control inputs, the robots move to new locations and update the position estimates of themselves and the targets, using newly collected measurements to perform an EKF update. 

\subsection{Objective Optimization by Greedy Algorithm}
\label{sec:optimization}


In the first stage, we focus on solving an unconstrained version of Program~\ref{opt1}, which is minimizing $\text{Tr}(\hat{\mathbf{P}})$. Even without constraints, the problem is still challenging due to the nonlinearity and nonconvexity of the objective function $\text{Tr}(\hat{\mathbf{P}}^t)$. To relax its hardness, we consider robots to have a discrete action space. In other words, each robot $i \in \mathcal{R}$ has a set of candidate control inputs $\mathcal{U}_{R_i}$, from which it can choose one at each time step. Taking advantage of the discrete action space, we design and present a greedy algorithm in Algorithm~\ref{algo1}\footnote{In Algorithm~\ref{algo1}, time step $t$ is dropped for the brevity of notation.} to select control input for each robot. 

Algorithm~\ref{algo1} takes as inputs the state estimates $(\hat{\bm{x}}_\mathcal{R}^{t-1},\hat{\mathbf{P}}_\mathcal{R}^{t-1})$, $(\hat{\bm{x}}_\mathcal{T}^{t-1},\hat{\mathbf{P}}_\mathcal{T}^{t-1})$ of robots and targets and outputs the robots' desired control inputs $\bm{{u}}_{\mathcal{R},\texttt{des}}^{t-1}$. It takes $N$ rounds to sequentially choose control inputs for all robots $\mathcal{R}$. For each robot $i$, its candidate control inputs are iterated through, such that the control input which yields the minimum $\text{Tr}(\hat{\mathbf{P}}^t)$ is selected as the ``desired" control input. $\text{Tr}_{i, \texttt{min}}$ (line 3) is initialized to be $+\infty$ and records the current minimum trace of the covariance matrix. Inside the inner for-loop (lines 4-10), each control input $\bm{u}_{R_i}^j \in \mathcal{U}_{R_i}$ is first used to propagate the dynamics of robot $i \in \mathcal{R}$ (line 5). This is the propagation step of the Extended Kalman Filter (EKF). Note that because Algorithm~\ref{algo1} runs in sequential order, the propagation step (line 5) applies to the exact robot under consideration, as well as those robots that have chosen their desired control inputs. However, for the robots which have not been iterated yet, they keep their previous state estimates, i.e., $\hat{\bm{x}}_{R_i}^{t|t-1}=\hat{\bm{x}}_{R_i}^{t-1}$. After that, we implement EKF's update step, denoted by the mapping $\bm{f}$, to generate posterior covariance matrices for the robots and targets $\hat{\mathbf{P}}_\mathcal{R}^{t|t}$ and $\hat{\mathbf{P}}_\mathcal{T}^{t|t}$ (line 6). Then, we compute the trace of the joint posterior covariance matrix of the robots and targets (line 7). If a lower uncertainty is found compared with the current minimum trace $\text{Tr}_{i, \texttt{min}}$, the latter is updated and the control input under consideration is set to be the desired control input (lines 8-10). In this way, Algorithm~\ref{algo1} returns the desired control inputs for all robots (line 11).

Notably, the greedy algorithm only takes two for-loops to iterate over the robots' candidate control inputs to select control inputs for all the robots. Therefore, it runs in polynomial time with time complexivity $O(\sum_{i=1}^N|\mathcal{U}_{R_i}|)$. In addition, the trace is a monotone function in the number of robots~\cite{jawaid2015submodularity}. That is, the larger the number of robots is, the lower the trace is. To optimize a monotone function, the greedy algorithm has been shown to provide a $1-c$ approximation of the optimal with $c\in[0,1]$ being the curvature of function (e.g., trace)~\cite{conforti1984submodular}. Thus, Algorithm~\ref{algo1} attains both fast running time and suboptimality guarantee to minimize the objective function  $\text{Tr}(\hat{\mathbf{P}})$. 


\renewcommand{\algorithmicrequire}{\textbf{Input:}}
\renewcommand{\algorithmicensure}{\textbf{Output:}}

\begin{algorithm}[t]
\caption{Greedy control input selection}
\begin{algorithmic}[1]\label{algo1}
\REQUIRE
   State estimates of robots $(\hat{\bm{x}}_\mathcal{R}^{t-1},\hat{\mathbf{P}}_\mathcal{R}^{t-1})$; 
   state estimates of the targets $(\hat{\bm{x}}_\mathcal{T}^{t-1},\hat{\mathbf{P}}_\mathcal{T}^{t-1})$;  
   Each robot $i$'s candidate set of control inputs $\mathcal{U}_{R_i}$. 

\ENSURE robot control actions $\bm{{u}}_{\mathcal{R},\texttt{des}}^{t-1}$.
\medskip 

\STATE $\hat{\bm{x}}_\mathcal{R}^{t|t-1} \leftarrow \hat{\bm{x}}_\mathcal{R}^{t-1}$;
\FOR{$i=1:N$}
\STATE $\text{Tr}_{i, \texttt{min}} \leftarrow +\infty$;
    \FOR{$\bm{u}_{R_i}^j \in \mathcal{U}_{R_i}$}
        \STATE $\hat{\bm{x}}_{R_i}^{t|t-1} \leftarrow \mathbf{A}_{R_i} \hat{\bm{x}}_{R_i}^{t-1} + \mathbf{B}_{R_i}\bm{u}_{R_i}^j$;
        
        \STATE $(\hat{\mathbf{P}}_{\mathcal{R}}^{t|t}, \hat{\mathbf{P}}_{\mathcal{T}}^{t|t}) \leftarrow \bm{f}\left(\hat{\bm{x}}_\mathcal{R}^{t|t-1}, \, \hat{\bm{x}}_\mathcal{T}^{t-1}, \, \hat{\mathbf{P}}_\mathcal{R}^{t-1}, \, \hat{\mathbf{P}}_\mathcal{T}^{t-1} \right)$;
        
        \STATE $\text{Tr}^{t|t} \leftarrow \mbox{\text{Tr}}\left(\hat{\mathbf{P}}_{\mathcal{R}}^{t|t}\right) + \mbox{\text{Tr}}\left(\hat{\mathbf{P}}_{\mathcal{T}}^{t|t} \right)$;
        \IF{$\text{Tr}^{t|t} < \text{Tr}_{i, \texttt{min}}$}
            \STATE $\text{Tr}_{i, \texttt{min}} \leftarrow \text{Tr}^{t|t}$;
            \STATE $\bm{u}_{R_i, \texttt{des}} \leftarrow \bm{u}_{R_i}^j$;
        \ENDIF 
    \ENDFOR
\ENDFOR
\STATE \textbf{return} $\bm{u}_{\mathcal{R}, \texttt{des}}$
\end{algorithmic}
\end{algorithm}

\subsection{Safety Constraints Guarantee by CBF}
Though the desired control inputs are derived efficiently by Algorithm~\ref{algo1}, the three constraints of Program~\ref{opt1} need to be ensured. First, the robots should remain connected with each other to maintain mutual information exchange. Second, the collision between robots should be avoided. Together, these constraints are termed safety constraints. Thirdly, robots have limited motion. These constraints are enforced in a Quadratic Program (QP) whose objective is to minimize the difference between the actual control inputs and the desired ones, with the connectivity maintenance and collision avoidance constraints formulated using CBFs. The CBFs ensure that the robots will always satisfy the safety constraints as long as their initial configuration satisfies the constraints. The QP also considers each robot's motion constraint. 

To maintain connectivity, first, the weighted adjacency matrix $\mathbf{A}^{t-1}=[a_{ij}^{t-1}]_{N\times N} \in \mathbb{R}^{N\times N}$ is computed based on the distance between each pair of robots as in the following:
\begin{align}
    d_{ij}^{t-1} &= \norm{\hat{\bm{x}}_{R_i}^{t-1} - \hat{\bm{x}}_{R_j}^{t-1}}
\end{align}
Adjacency between robot $i$ and robot $j$ is computed by the following:
\begin{equation}
a_{ij}^{t-1} = \left\{
        \begin{array}{ll}
            e^\frac{{\left(r_c^2 - {d_{ij}^{t-1}}^2\right)^2}}{\sigma} - 1 & \quad \mbox{if }d_{ij}^{t-1} \leq r_c \\
            0 & \quad \mbox{otherwise}
        \end{array}
    \right.
\end{equation}
where $\sigma$ is a normalization parameter to keep values between 0 and 1. Using the diagonalized form of $\mathbf{A}^{t-1}$, termed $\mathbf{D}^{t-1}$, the second smallest eigenvalue $\lambda_2$ of the adjacency Laplacian matrix $\mathbf{L}^{t-1}$ along with its corresponding eigenvector $\bm{v}_2^{t-1}$ is computed as:

\begin{align}
    \mathbf{D}^{t-1} &= \text{diag}\left(\mathbf{A}^{t-1}\right) \\
    \mathbf{L}^{t-1} &= \mathbf{D}^{t-1} - \mathbf{A}^{t-1} \\
    \lambda_2^{t-1} &= \lambda_2\left(\mathbf{L}^{t-1}\right) \\
    \bm{v}^{t-1}_2 &= \text{eigenvector}_2\left(\mathbf{L}^{t-1}\right)
\end{align}
Recall that $\lambda_2$ serves as a scalar metric for quantifying the network connectivity of the robot team. Specifically, if $\lambda_2$ is greater than a small positive number $\epsilon$, here selected as $10^{-5}$, then the team is connected and can exchange information. With this metric, the QP is formulated as:


\begin{subequations}
        \label{eq:qp}
        \begin{align}
        \min_{\bm{u}^{t-1}} \quad & \frac{1}{2} \left\lVert \bm{u}_{\mathcal{R}}^{t-1} - \bm{u}_{\mathcal{R},\texttt{des}}^{t-1} \right\rVert^2 \label{eq:qp_cost}\\ 
        \textrm{s.t.} \quad 
        & -\bm{\beta}^{t-1}\bm{u}_{\mathcal{R}}^{t-1} \leq \lambda_2 - \epsilon, \label{eq:qp_connect}\\
        & d_{ij}^{t-1}\left(\bm{u}_{R_i}^{t-1} - \bm{u}_{R_j}^{t-1}\right)^T + \frac{1}{2}\left(\norm{d_{ij}^{t-1}}^2 - d_{\min}^2\right)^3 \geq \bm{0}, \nonumber\\
        & \forall i\neq j, i,j\in \mathcal{R}  \label{eq:qp_collision_avoid} \\
        & \|\bm{u}_{R_i}^{t-1}\|_2 < u_{\max},  ~\forall i \in \mathcal{R}.
        \end{align}
\end{subequations}
where $\bm{\beta}^{t-1}$ represents the derivative of $\lambda_2$ with respect to each robot's state, and is computed as:

\begin{equation}
    \bm{\beta}_{i}^{t-1} = \frac{\partial\lambda_2^{t-1}}{\partial \bm{x}^t_i} = \sum_{j = 1}^{n} \frac{\partial a_{ij}^{t-1}}{\partial \bm{x}_i^{t-1}}\left({\bm{v}^{t-1}_{2, i}} - {\bm{v}_{2, j}^{t-1}}\right)^2, \forall i \in \mathcal{R}.
\end{equation}
Note that Eq.~\ref{eq:qp_collision_avoid} is the collision avoidance constraint also derived using a CBF. The solution to the QP gives the real control inputs to be executed by the robot team. 

\subsection{Target Estimates by EKF}
Target state estimates are obtained by recursively running the Extended Kalman Filter (EKF).
At each time step $t$, prior target knowledge is computed by propagating posterior information from the previous time-step $t-1$:
\begin{align} \label{eq:target_priors}
    & \bm{\bar{x}}^{t}_{T_{i}} = \mathbf{A}_{T_i}\hat{\bm{x}}_{T_i}^{t-1}, \\
    & \mathbf{\bar{P}}^{t}_{T_{i}} = \mathbf{A}_{T_i}\hat{\mathbf{P}}^{t-1}_{T_i}\mathbf{A}_{T_i}^T + \mathbf{G}_{T_i}\mathbf{Q}_{d,T_i}^{t-1}\mathbf{G}_{T_i}^T
\end{align}
where $\bm{\bar{x}}^{t}_{T_{i}}$ is the prior estimate of the state of target $i$ at time $t$ and $\hat{\bm{x}}^{t-1}_{T_{i}}$ is the posterior estimate of the state of target $i$ at time $t-1$. $\hat{\mathbf{P}}^{t-1}_{T_i}$ is the posterior covariance of estimate for target $i$ at time $t-1$. $\mathbf{\bar{P}}^{t}_{T_{i}}$ denotes the prior covariance of the estimate at time $t$ and has an added noise component based on the white-noise covariance $\mathbf{Q}_{d,T_i}^{t-1}$. Target state estimates are vertically composed into a single matrix by:
\begin{equation}
    \bm{\bar{x}}^{t}_\mathcal{T} = \left[\left(\bm{\bar{x}}^{t}_{T_1}\right)^T, \mbox{ }\left(\bm{\bar{x}}^{t}_{T_2}\right)^T \mbox{ },\cdots,\mbox{ } \left(\bm{\bar{x}}^{t}_{T_M}\right)^T\right]^T
    \label{eq:target_state_composition}
\end{equation}
Target covariance estimates are composed into a single block-diagonal matrix:
\begin{equation}
    \mathbf{\bar{P}}^{t}_\mathcal{T} = \mbox{blockdiag}\left(\mathbf{\bar{P}}^{t}_{T_1}, \mbox{ }\mathbf{\bar{P}}^{t}_{T_2} \mbox{ },\cdots,\mbox{ } \mathbf{\bar{P}}^{t}_{T_M}\right)
    \label{eq:target_cov_composition}
\end{equation}
At each time step, each robot, here outfitted with both linear and GPS sensors, observes every target and every robot. Measurements are composed of the following:
\begin{align}
    &\bm{z}_{T_j}^t = \left[\left(\bm{z}_{R_1, T_j}^t\right)^T, \cdots, \left(\bm{z}_{R_N, T_j}^t\right)^T\right]^T, \\
    &\bm{z}_\mathcal{T}^t = \left[\left(\bm{z}_{T_1}^t\right)^T, \cdots, \left(\bm{z}_{T_{M}}^t\right)^T\right]^T 
\end{align}
Estimated measurements are determined by simulating measurements based on prior knowledge:
\begin{equation}
    \bm{\bar{z}}_{R_i, T_j}^{t} = \mathbf{h}\left(\bm{\bar{x}}_{R_i}^{t}, \bm{\bar{x}}_{T_j}^t\right)
    \label{eq:simulated_measurements}
\end{equation}
The measurement error is then found by:
\begin{align}\label{eq:measurement_error}
    \tilde{\bm{z}}_{R_i, T_j}^{t} &= \bm{z}_{R_i, T_j}^{t} - \bm{\bar{z}}_{R_i, T_j}^t \\
    &\approx \mathbf{H}_{R_i, T_j}^t\tilde{\bm{x}}_{R_i}^t + \mathbf{H}_{T_j, R_i}^t\tilde{\bm{x}}_{T_j}^t + \bm{n}_{R_i, T_j}^t
\end{align}
where $\tilde{\bm{z}}_{R_i, T_j}$ is the measurement error, $\tilde{\bm{x}}_{R_i}^t$ and $\tilde{\bm{x}}_{T_j}^t$ are the estimation errors for robot $i$ and target $j$, and $ \mathbf{H}_{R_i, T_j}^t$ and $\mathbf{H}_{T_j, R_i}^t$ are the linearizations for measurement function $\mathbf{h}$ centered around $\bm{\bar{x}}_{R_i}^t$ and $\bm{\bar{x}}_{T_j}^t$, respectively. In ensemble form, the measurement matrix $\mathbf{H}_\mathcal{T}^{t}$ after linearization of all targets by all robots is:
\begin{align}\label{eq:compose_linearizations}
    &\mathbf{H}_{T_j}^t = \left[\left(\mathbf{H}_{T_j, R_1}^t\right)^T, \cdots, \left(\mathbf{H}_{T_j, R_N}^t\right)^T\right]^T, \\
    &\mathbf{H}_\mathcal{T}^t = \left[\left(\mathbf{H}_{T_1}^t\right)^T, \cdots, \left(\mathbf{H}_{T_M}^t\right)^T\right]^T
\end{align}
Using the target measurement matrix, the update stage of EKF produces target posterior estimates:
\begin{align}\label{eq:target_kalman_update}
    & \mathbf{S}^{t}_\mathcal{T} = \mathbf{H}^{t}_\mathcal{T}\mathbf{\bar{P}}^{t}_\mathcal{T}(\mathbf{H}^{t}_\mathcal{T})^T + \mathbf{R}^{t}_\mathcal{T} \\
    & \mathbf{K}^{t}_\mathcal{T} = \mathbf{\bar{P}}^{t}_\mathcal{T}(\mathbf{H}^{t}_\mathcal{T})^T(\mathbf{S}^{t}_\mathcal{T})^{-1} \\
    & \mathbf{\bar{K}}^{t}_\mathcal{T} = \mathbf{I} - \mathbf{K}^{t}_\mathcal{T}\mathbf{H}^{t}_\mathcal{T} \\
    & \hat{\mathbf{P}}^{t}_\mathcal{T} = \mathbf{\bar{K}}^{t}_\mathcal{T}\mathbf{\bar{P}}^{t}_\mathcal{T}(\mathbf{\bar{K}}^{t}_\mathcal{T})^T + \mathbf{K}^{t}_\mathcal{T}\mathbf{R}^{t}_\mathcal{T}(\mathbf{K}^{t}_\mathcal{T})^T \\
    & \hat{\bm{x}}^{t}_\mathcal{T} = \bm{\bar{x}}^{t}_\mathcal{T} + \mathbf{K}^{t}_\mathcal{T}\tilde{\bm{z}}^{t}_\mathcal{T}
\end{align}
\subsection{Localization Estimates by EKF}
Similar to the estimation of target positions, the localization of robots is also an iterative state estimation problem, and could thus be addressed using EKF. During the prediction stage at time step $t$, prior robot estimates are computed as:  
\begin{align} \label{prior_robot}
    & \bm{\bar{x}}^{t}_{R_{i}} = \mathbf{A}_{R_i}\hat{\bm{x}}_{R_i}^{t-1} + \mathbf{B}_{R_i}\bm{u}_{R_i}^{t-1}\\
    & \mathbf{\bar{P}}^{t}_{R_{i}} = \mathbf{A}_{R_i}\hat{\mathbf{P}}^{t-1}_{R_i}\mathbf{A}_{R_i}^T +  \mathbf{G}_{R_i}\mathbf{Q}_{d,R_i}^{t-1}\mathbf{G}_{R_i}^T
\end{align}
\ie propagating posterior estimates from the previous time step. In matrix form, the prior estimates and the corresponding prior covariance for the whole robot team are:
\begin{align}\label{eq:robot_prior_composition}
    &\bm{\bar{x}}^{t}_\mathcal{R} = \left[\left(\bm{\bar{x}}^{t}_{R_1}\right)^T, \mbox{ }\left(\bm{\bar{x}}^{t}_{R_2}\right)^T \mbox{ },\cdots,\mbox{ } \left(\bm{\bar{x}}^{t}_{R_N}\right)^T\right]^T \\
    &\mathbf{\bar{P}}^{t}_\mathcal{R} = \mbox{blockdiag}\left(\mathbf{\bar{P}}^{t}_{R_1}, \mbox{ }\mathbf{\bar{P}}^{t}_{R_2} \mbox{ },\cdots,\mbox{ } \mathbf{\bar{P}}^{t}_{R_N}\right)
\end{align}
Note that each robot is equipped with one range and one bearing sensor, both of whose measurements are essentially non-linear in terms of the real states of objects they are sensing. Under this assumption, the measurement model is linearized for the update stage of EKF which derives the posterior estimates. First, to obtain the measurement error, the simulated measurement of robot $j$ by robot $i$ is computed as follows using the prior estimates:
\begin{equation}
    \bm{\bar{z}}_{R_i, R_j}^{t} = \mathbf{h}\left(\bm{\bar{x}}_{R_i}^{t}, \bm{\bar{x}}_{R_j}^t\right)
    \label{eq:robot_simulated_measurements}
\end{equation}
Each robot computes a measurement of itself using a GPS sensor. To simplify measurement composition, the GPS measurement is notated as $\bm{z}_{R_i, R_i}^t = \bm{z}_{R_i, GPS}^t$. Robots measure each other by relying on linear sensor suites. We denote robot $i$'s measurement of robot $j$ as $\bm{z}_{R_i, R_j}^t$, robot $i$'s measurement of all robots as $\bm{z}_{R_i}^{t}$, and all-to-all measurements of the robot team as $z_\mathcal{R}^t$. The latter two are computed in Eq.~\ref{eqn:z_R_i} and~\ref{eqn:z_R}, respectively.  
\begin{align}
    &\bm{z}_{R_{i}}^t = \left[\left(\bm{z}_{R_i, R_1}^t\right)^T, \cdots, \left(\bm{z}_{R_i, R_N}^t\right)^T\right]^T, \label{eqn:z_R_i}\\
    &\bm{z}_\mathcal{R}^t = \left[\left(\bm{z}_{R_{1}}^t\right)^T, \cdots, \left(\bm{z}_{R_N}^t\right)^T\right]^T \label{eqn:z_R}
\end{align}
Using the real and simulated measurements, the measurement error for robot $i$ measuring robot $j$ is defined as:
\begin{align}\label{eq:robot_measurement_error}
    \tilde{\bm{z}}_{R_i, R_j}^{t} &= \bm{z}_{R_i, R_j}^{t} - \bm{\bar{z}}_{R_i, R_j}^t \\
    &\approx \mathbf{H}_{R_i, R_j}^t\tilde{\bm{x}}_{R_i}^t + \mathbf{H}_{R_j, R_i}^t\tilde{\bm{x}}_{R_j}^t + \bm{n}_{R_i, R_j}^t
\end{align}
where $\mathbf{H}_{R_i, R_j}^t$ and $\mathbf{H}_{R_j, R_i}^t$ are the Jacobian matrices for measurement model $\mathbf{h}$ instantiated at $\bm{\bar{x}}_{R_i}^t$ and $\bm{\bar{x}}_{R_j}^t$, respectively. The measurement error matrix for the whole robot team $\tilde{\bm{z}}^{t}_\mathcal{R}$ could be composed in similar fashion to Eq.~\ref{eqn:z_R_i}-\ref{eqn:z_R}. Each absolute GPS measurement is modeled as $\bm{z}_{R_i, R_i}^t = \bm{x}_{R_i}^t + \bm{n}_{R_i, R_i}^t$. With linearization, its Jacobian is simply $\mathbf{H}_{R_i, R_i}^t = \mathbf{I}$. 
The robot measurement matrix $\mathbf{H}_\mathcal{R}$ for all robots is therefore ensembled as: 

\begin{align}
    \begin{split}
        \mathbf{H}_{R_j}^t =  &\left[\left(\mathbf{H}_{R_j, R_1}^t\right)^T, \cdots, \left(\mathbf{H}_{R_j, R_{j-1}}^t\right)^T,\right. \\ 
        &\left.  \left(\mathbf{H}_{R_j, R_j}^t\right)^T, \cdots, \left(\mathbf{H}_{R_j, R_N}^t\right)^T\right]^T, 
    \end{split}
\end{align}

\begin{equation}
    \mathbf{H}_\mathcal{R}^t = \left[\left(\mathbf{H}_{R_1}^t\right)^T, \cdots, \left(\mathbf{H}_{R_N}^t\right)^T\right]^T
\end{equation}
With the robot measurement matrix, the EKF update for localization is performed to produce posterior estimates of the robots:
\begin{align}\label{eq:robot_kalman_update}
    & \mathbf{S}^{t}_\mathcal{R} = \mathbf{H}^{t}_\mathcal{R}\mathbf{\bar{P}}^{t}_\mathcal{R}(\mathbf{H}^{t}_\mathcal{R})^T + \mathbf{R}^{t}_\mathcal{R} \\
    & \mathbf{K}^{t}_\mathcal{R} = \mathbf{\bar{P}}^{t}_\mathcal{R}(\mathbf{H}^{t}_\mathcal{R})^T(\mathbf{S}^{t}_\mathcal{R})^{-1} \\
    & \mathbf{\bar{K}}^{t}_\mathcal{R} = \mathbf{I} - \mathbf{K}^{t}_\mathcal{R}\mathbf{H}^{t}_\mathcal{R} \\
    & \hat{\mathbf{P}}^{t}_\mathcal{R} = \mathbf{\bar{K}}^{t}_\mathcal{R}\mathbf{\bar{P}}^{t}_\mathcal{R}(\mathbf{\bar{K}}^{t}_\mathcal{R})^T + \mathbf{K}^{t}_\mathcal{R}\mathbf{R}^{t}_\mathcal{R}(\mathbf{K}^{t}_\mathcal{R})^T \\
    & \hat{\bm{x}}^{t}_\mathcal{R} = \bm{\bar{x}}^{t}_\mathcal{R} + \mathbf{K}^{t}_\mathcal{R}\tilde{\bm{z}}^{t}_\mathcal{R}
\end{align}

\section{Simulation Results}
\label{sec:results}
\subsection{Qualitative Results}

\begin{figure*}[th!]
\centering{
\subfigure[$N=4, M=5$\label{fig:5i_4d_gazebo}]{\includegraphics[width=0.5\columnwidth]{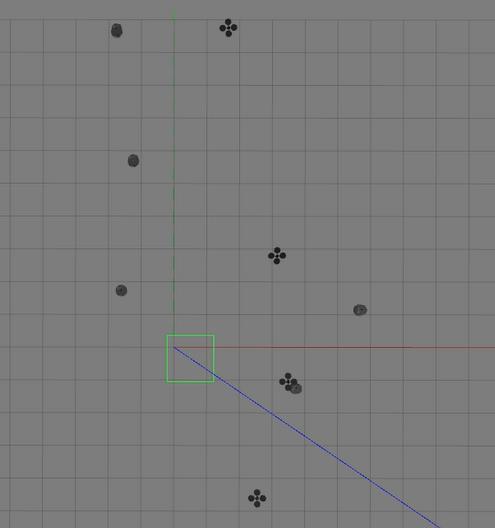}}
\hspace{6mm}\subfigure[$N=5, M=5$\label{fig:5i_5d_gazebo}]{\includegraphics[width=0.5\columnwidth]{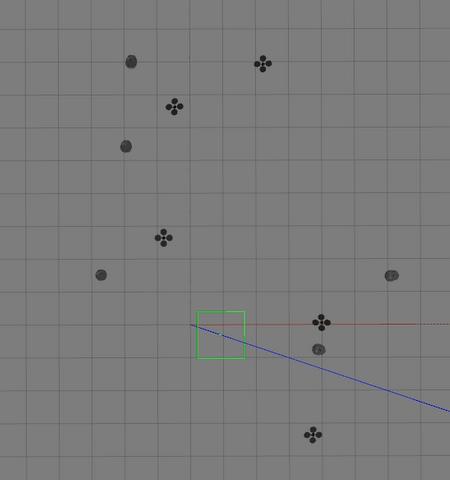}}
\hspace{6mm}\subfigure[$N=6, M=5$\label{fig:5i_6d_gazebo}]{\includegraphics[width=0.5\columnwidth]{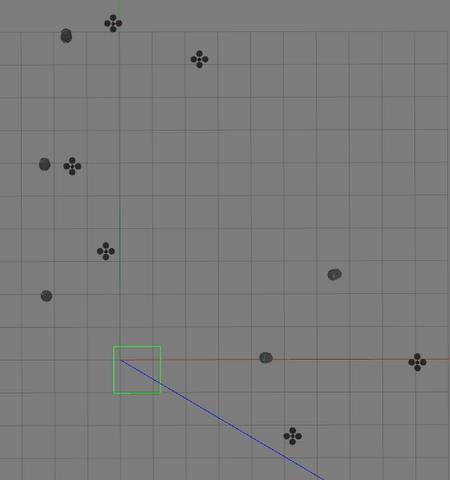}}
\caption{Gazebo screenshots for 3 different settings (4 robots track 5 targets for~\ref{fig:5i_4d_gazebo}, 5 robots track 5 targets for~\ref{fig:5i_5d_gazebo}, and 6 robots track 5 targets for~\ref{fig:5i_6d_gazebo}), all taken at time step $t = 50$.}
\label{fig:gazebo}
}
\end{figure*}

Simulated experiments were run using Gazebo Simulator on a Ubuntu 20.04 Desktop to qualitatively demonstrate the performance of the proposed framework. Inside the Gazebo environment, targets are Scarab cars running on the ground, while the robots are HummingBird drones. Three different scenarios are considered: 4 drones track 5 targets, 5 drones track 5 targets, and 6 drones track 5 targets, to verify the effectiveness of the proposed framework under various settings (i.e., varying number of robots $N$ and varying number of targets $M$). A demo of the Gazebo experiments is available online\footnote{\url{https://youtu.be/H7gBGyEb0Lw}}.

The parameters used in the simulations were chosen through trial and error. For all three settings, the maximum velocity of drones is $1.0\ m/s$, the safety distance is $d_\texttt{min} = 3.0\ m$, the initial covariance for each target is $1000\mathbf{I}_2$, and the initial covariance for each robot is $25\mathbf{I}_2$\footnote{We only consider the planar positions of drones.}. When there are 6 or 5 robots tracking 5 targets, the maximum communication range is $r_c = 10.0$, while it reduces to $r_c=8.0$ when only 4 robots are available. The selection of such a configuration originates from the fact that increasing the team size of robots makes it easier to obtain accurate estimates, but decreasing their communication range could make the task more challenging and thus leads to more convincing evidence. 

Gazebo screenshots for all three settings taken at time step $t = 50$ are shown in parallel in Fig.~\ref{fig:gazebo}. Note that the drones take off from near the origin. As observed, the robots closely follow the Scarab cars for a more accurate estimation of the targets' positions. Moreover, under each of the three settings, the robots form sub-teams automatically to adapt to the distribution of targets in the environment. For instance, in the 5-robot-5-target case, 3 targets move to the upper left region, while 2 targets move to the lower right. Correspondingly, the 5 robots divide themselves into one sub-team of 3 members and the other sub-team consisting of the rest 2 members. The former focuses on the 3 targets in the upper left, while the latter takes the responsibility of tracking the other two targets on the lower right. This is shown in Fig.~\ref{fig:5i_5d_gazebo}. Similar divisions of labor can be seen in Fig.~\ref{fig:5i_4d_gazebo} and Fig.~\ref{fig:5i_6d_gazebo} for the other two settings.

\begin{figure}
\centering
\captionsetup[subfigure]{font=scriptsize,labelfont=scriptsize}
    \subfigure[$N=4, M=5$]{
      \includegraphics[width=0.296\columnwidth]{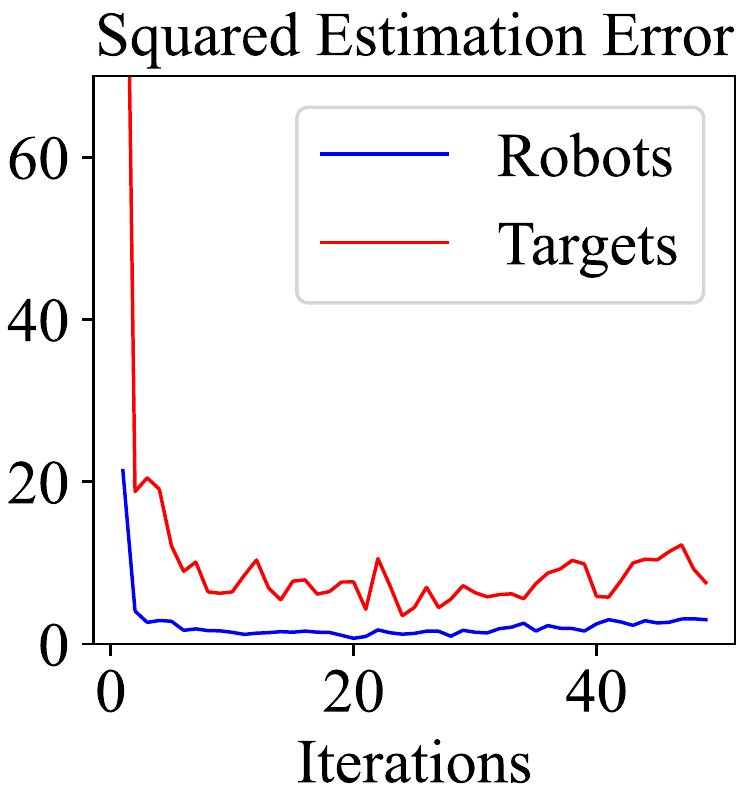}
      \label{fig:5i_4d_sqr}
    }
    \subfigure[$N=5, M=5$]{
      \includegraphics[width=0.296\columnwidth]{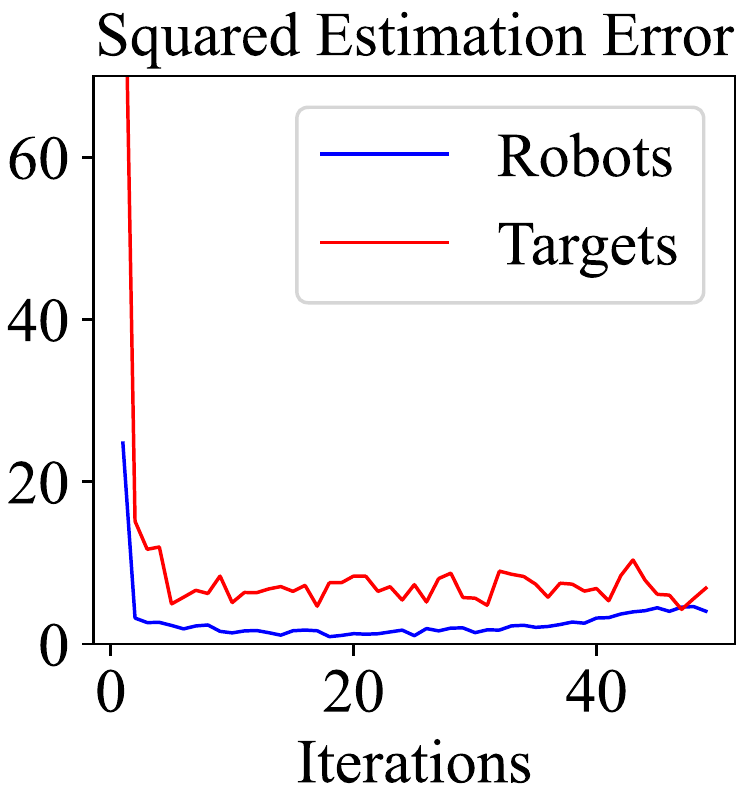}
      \label{fig:5i_5d_sqr}
    }
    \subfigure[$N=6, M=5$]{
      \includegraphics[width=0.296\columnwidth]{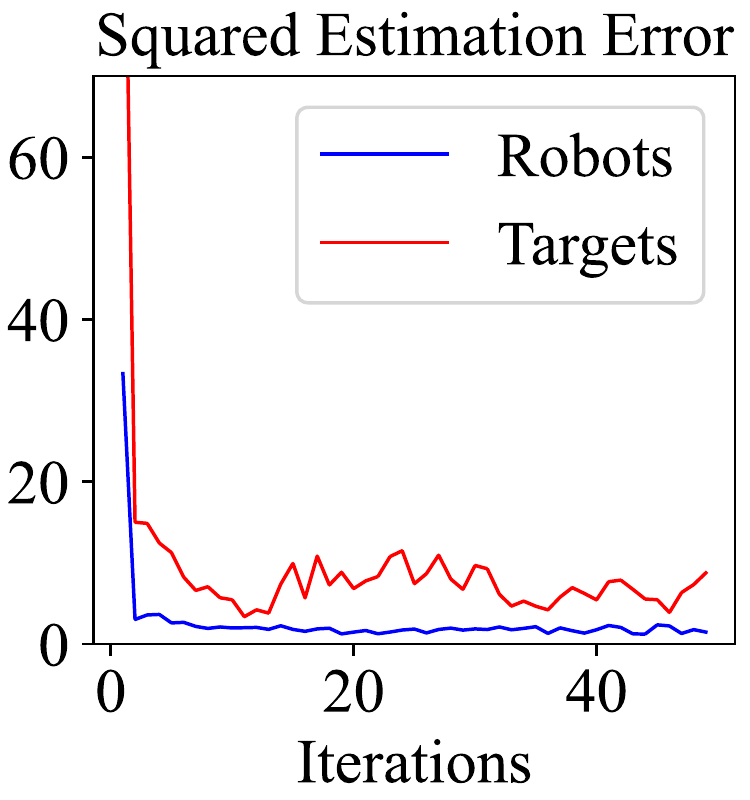}
      \label{fig:5i_6d_sqr}
    }
    \quad 
    \subfigure[$N=4, M=5$]{
      \includegraphics[width=0.296\columnwidth]{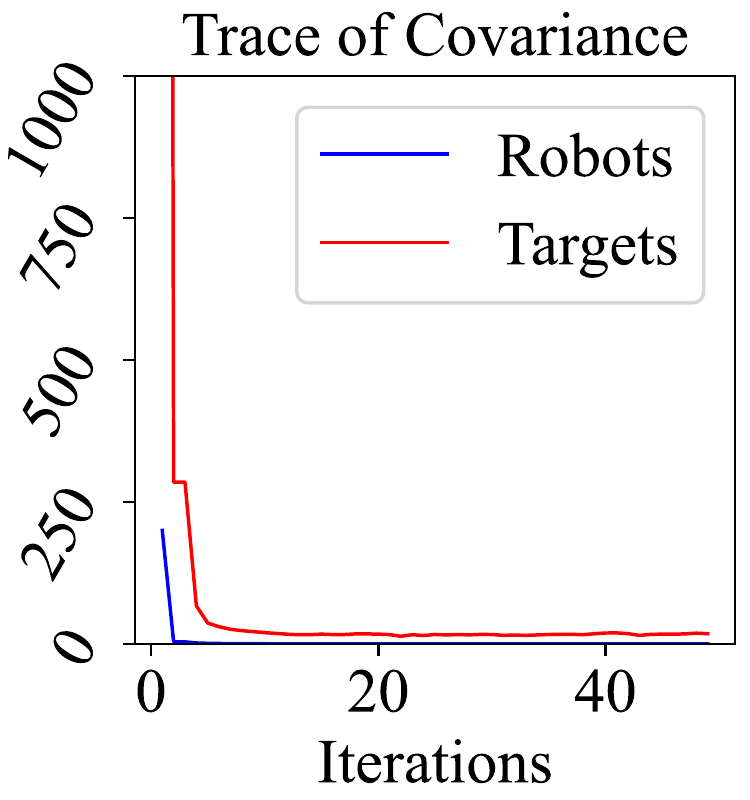}
      \label{fig:5i_4d_trace}
    }
    \subfigure[$N=5, M=5$]{
      \includegraphics[width=0.296\columnwidth]{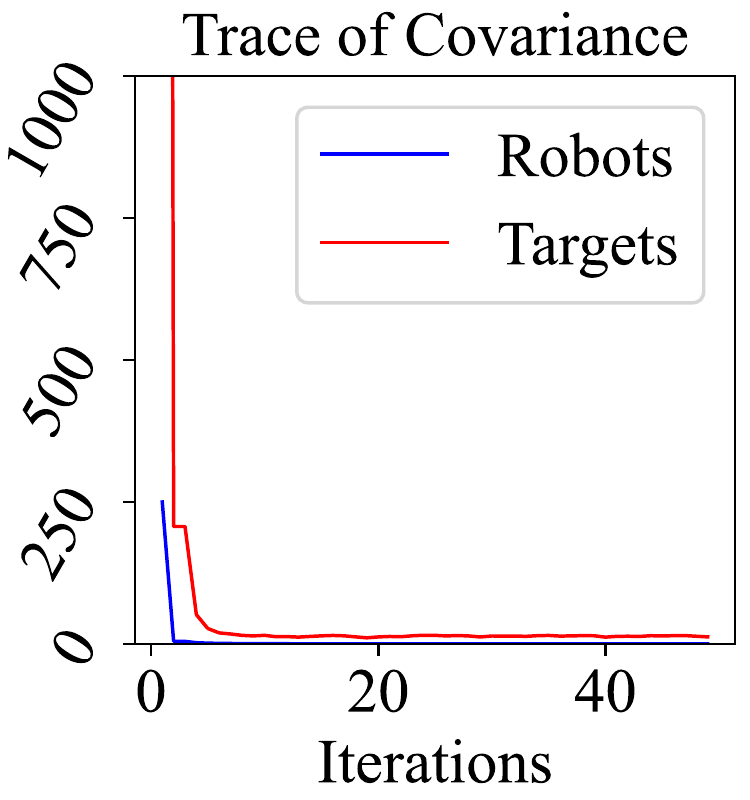}
      \label{fig:5i_5d_trace}
    }
    \subfigure[$N=6, M=5$]{
      \includegraphics[width=0.296\columnwidth]{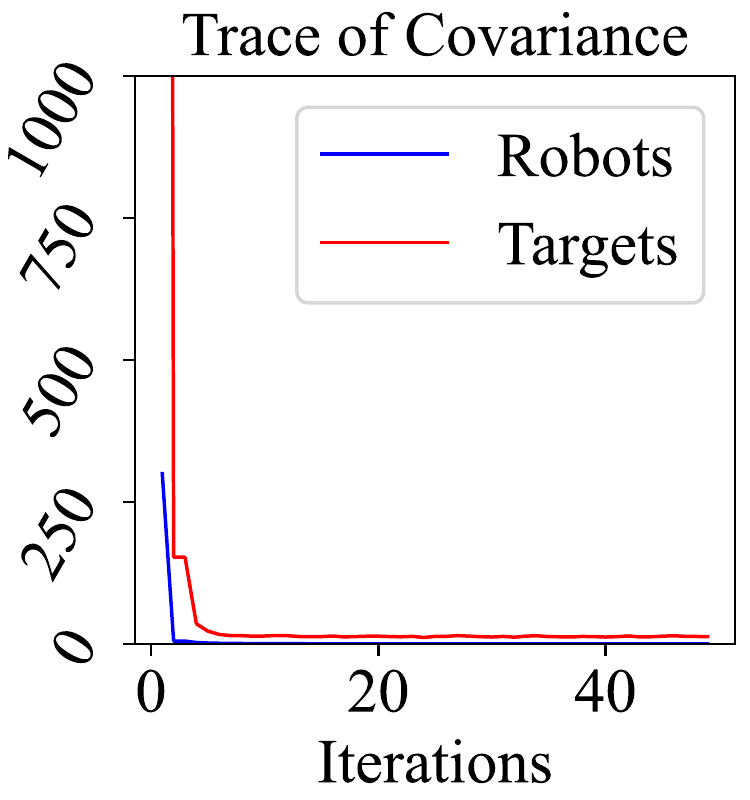}
      \label{fig:5i_6d_trace}
    }
    \quad 
    \subfigure[$N=4, M=5$]{
      \includegraphics[width=0.296\columnwidth]{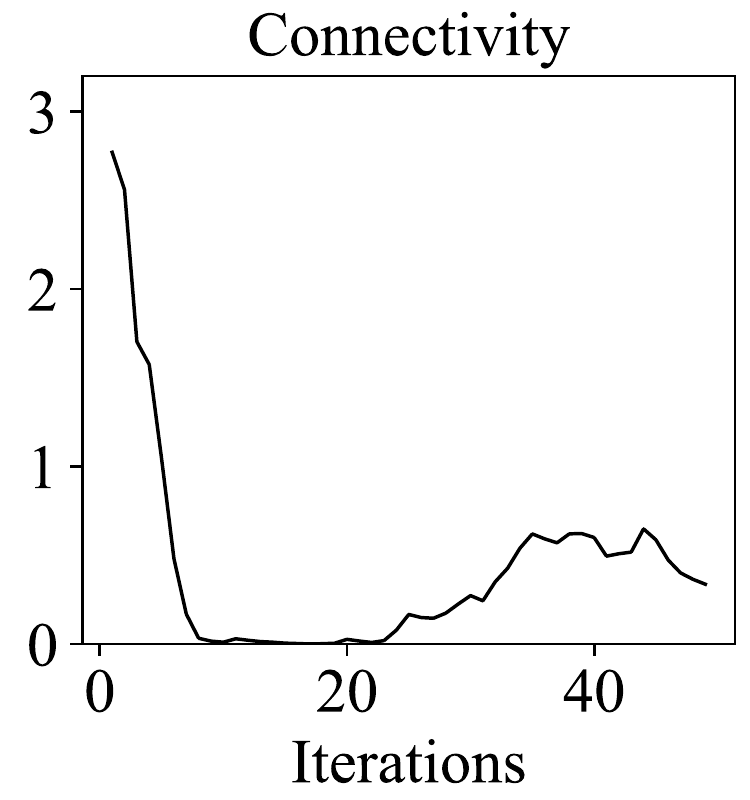}
      \label{fig:5i_4d_conn}
    }
    \subfigure[$N=5, M=5$]{
      \includegraphics[width=0.296\columnwidth]{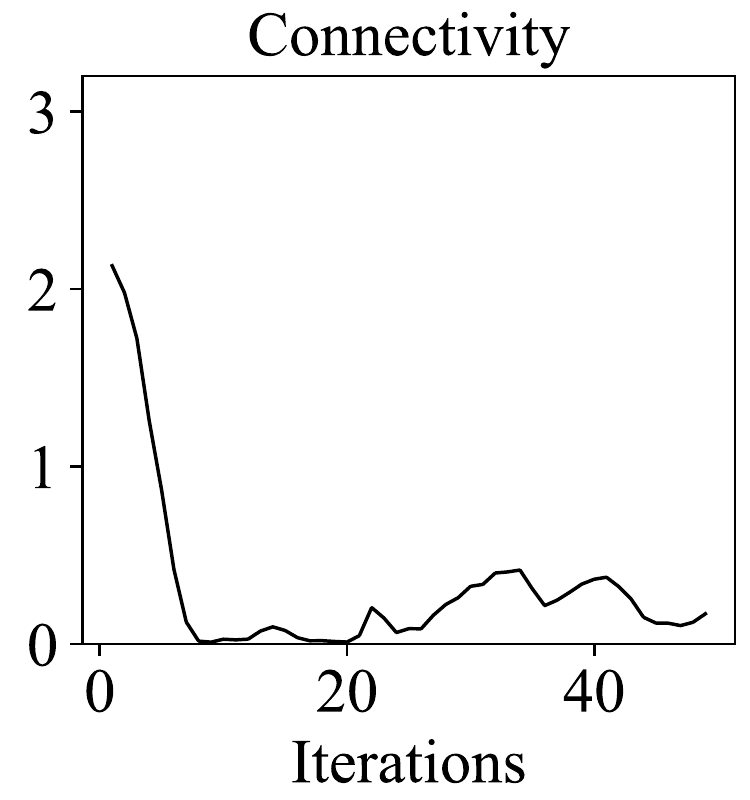}
      \label{fig:5i_5d_conn}
    }
    \subfigure[$N=6, M=5$]{
      \includegraphics[width=0.296\columnwidth]{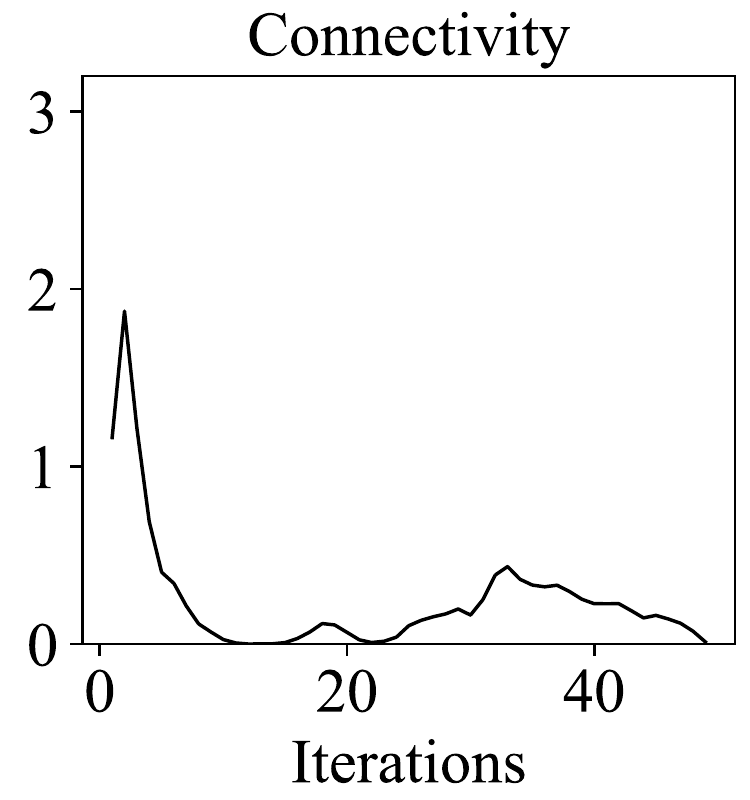}
      \label{fig:5i_6d_conn}
    }
    \quad 
    \subfigure[$N=4, M=5$]{
      \includegraphics[width=0.296\columnwidth]{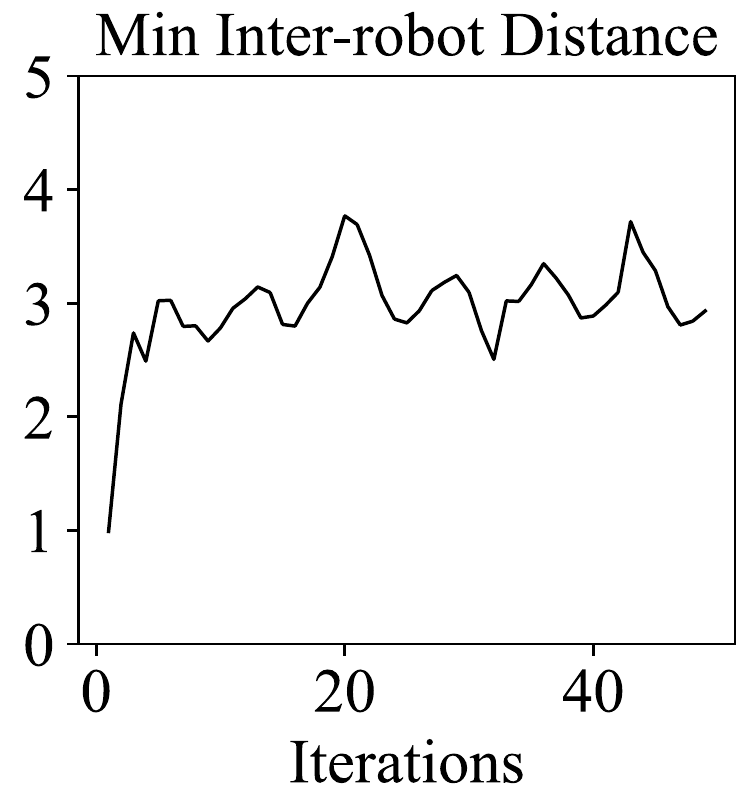}
      \label{fig:5i_4d_dist}
    }
    \subfigure[$N=5, M=5$]{
      \includegraphics[width=0.296\columnwidth]{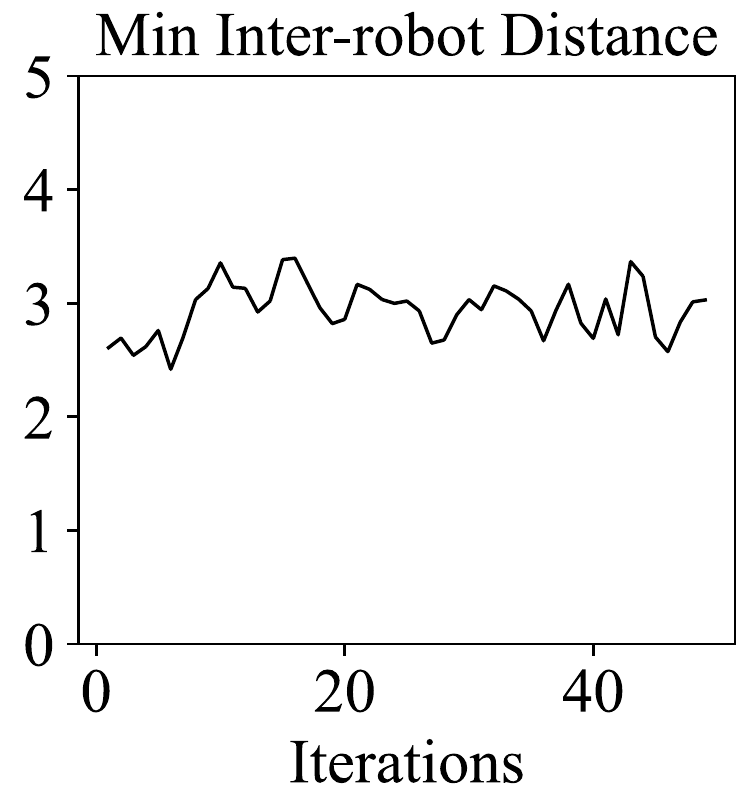}
      \label{fig:5i_5d_dist}
    }
    \subfigure[$N=6, M=5$]{
      \includegraphics[width=0.296\columnwidth]{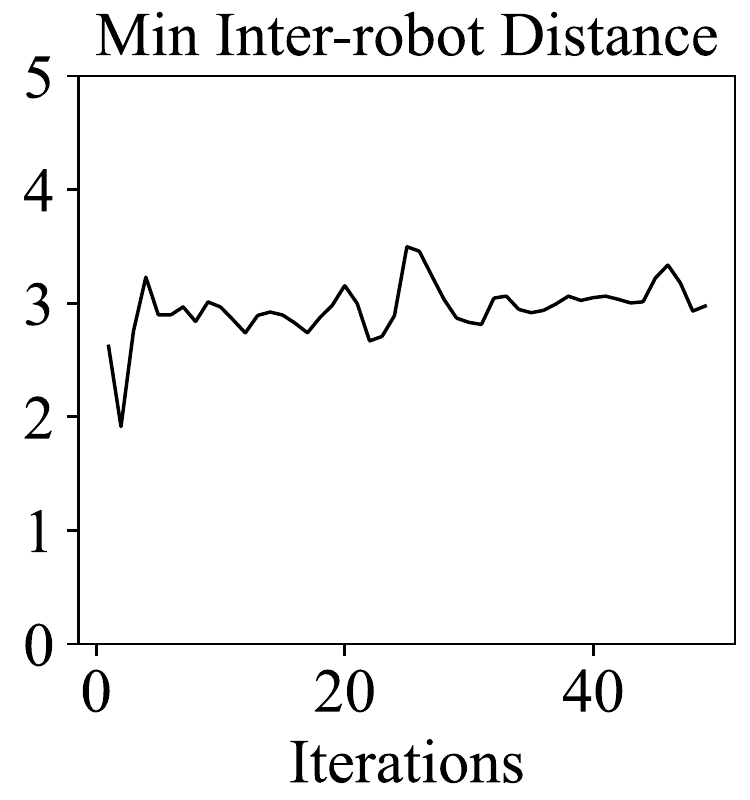}
      \label{fig:5i_6d_dist}
    }
    \caption{Illustration of qualitative results for three different settings: 4 robots track 5 targets (column 1), 5 robots track 5 targets (column 2), and 6 robots track 5 targets (column 3). Sub-figures~\ref{fig:5i_4d_sqr},~\ref{fig:5i_5d_sqr}, and~\ref{fig:5i_6d_sqr} show the respective squared estimation errors for both localization and target tracking of the three cases. Fig.\ref{fig:5i_4d_trace},~\ref{fig:5i_5d_trace}, and~\ref{fig:5i_6d_trace} show the respective weighted trace of the estimated covariance matrices. The third row, \ie Fig.~\ref{fig:5i_4d_conn}, ~\ref{fig:5i_5d_conn}, and ~\ref{fig:5i_6d_conn}, represents the network connectivity throughout 50 time steps. Fig.~\ref{fig:5i_4d_dist}, ~\ref{fig:5i_5d_dist}, and ~\ref{fig:5i_6d_dist} show the estimated minimum inter-robot distance. }
        \label{fig:qualitative_results}
\end{figure}

A comprehensive evaluation of the proposed framework includes the following two aspects: (i) accuracy of the joint localization and tracking task, and (ii) effectiveness of the CBF constraints, \ie connectivity maintenance and collision avoidance. The two metrics adopted for (i) are the squared estimation error and the sum of the traces of the covariance matrices, for both localization and target tracking. The squared estimation error for localization at time step $t$ is defined as:
\begin{equation}
    \sum_{i=1}^{N}\| \bm{x}_{R_i}^{t} - \hat{\bm{x}}_{R_i}^{t}\|^2,
\end{equation}
and similarly for target tracking as:
\begin{equation}
    \sum_{j=1}^{M}\| \bm{x}_{T_j}^{t} - \hat{\bm{x}}_{T_j}^{t}\|^2.
\end{equation}
As for (ii), the validity of the CBF constraint for connectivity maintenance and inter-robot collision avoidance is demonstrated through the algebraic connectivity $\lambda_2$ of the communication graph and the minimum inter-robot distance, respectively. These results are shown in Fig.~\ref{fig:qualitative_results}.


\begin{figure}
\centering{
\subfigure[Trace of the minimal estimated covariance for each algorithm's selected control inputs\label{fig:quant_cov_trace}]{\includegraphics[width=0.7\columnwidth]{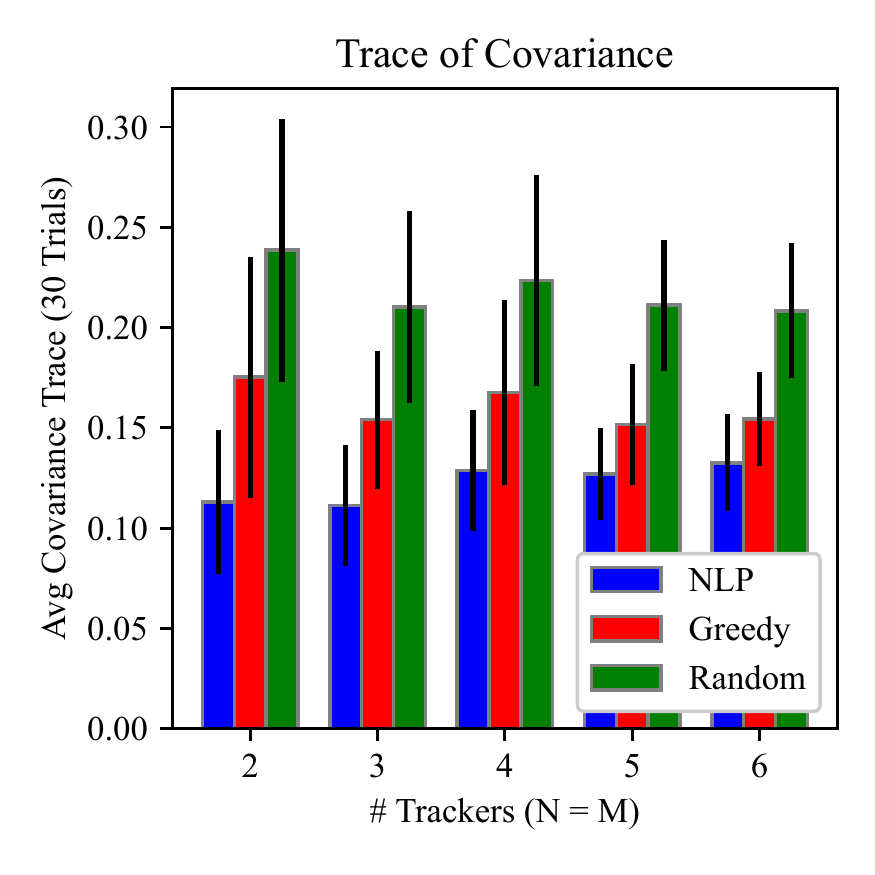}}
\subfigure[Computation time\label{fig:quant_comp_time}]{\includegraphics[width=0.7\columnwidth]{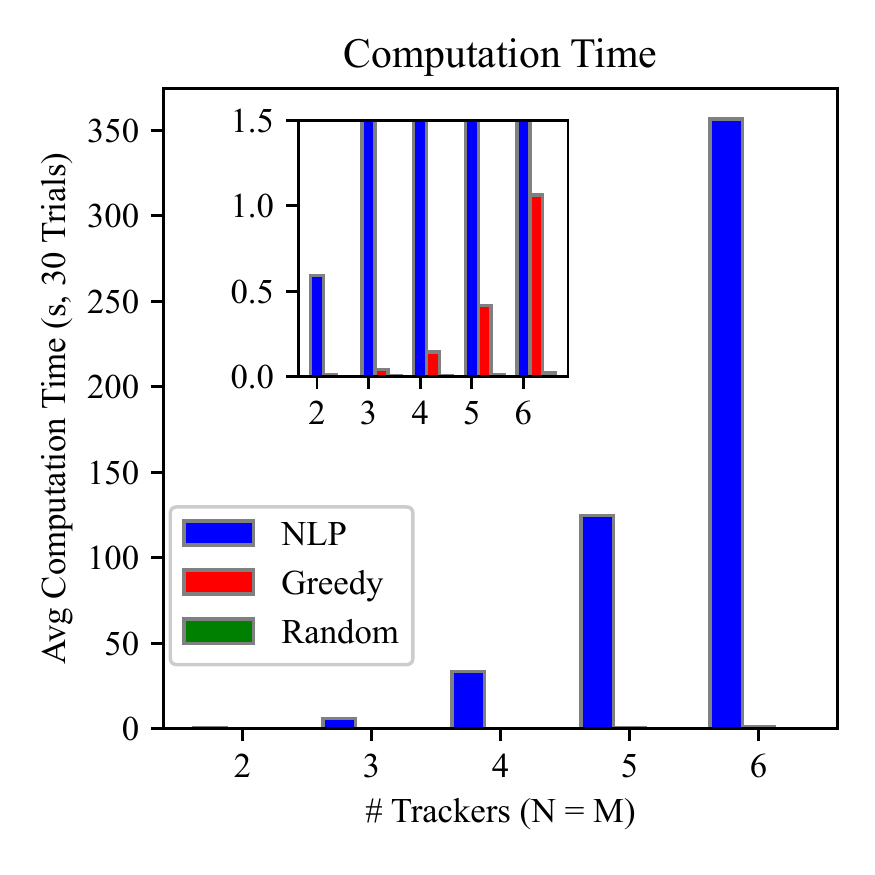}}
\caption{Single-step algorithm results with plotted standard deviations averaged over 30 randomized trials for a range of N=M settings.}
\label{fig:quanti-5i-4d}
}
\end{figure}

The four rows of Fig.~\ref{fig:qualitative_results} correspond to the squared estimation error, the trace of covariance, the algebraic connectivity $\lambda_2$, and the minimum inter-robot distance, respectively. Each column corresponds to one of the three settings described above. From the first row, it can be noted that while the given initial prior knowledge about the states of robots and targets is quite poor, as reflected by the large initial error, the squared estimation error quickly drops as the system adaptively takes actions to reconfigure the positions of robots. Throughout the whole process of task execution, the squared error is kept at a small level. Similar observations hold for the trace of covariance as shown in Fig.~\ref{fig:5i_4d_trace},~\ref{fig:5i_5d_trace}, and~\ref{fig:5i_6d_trace}. Combining these two lines of evidence, the proposed system's ability to achieve accurate joint localization and tracking is verified. 

Besides the metrics on the accuracy, results in row 3 and row 4 of Fig.~\ref{fig:qualitative_results} validate that the CBF constraints are effective. In row 3, the algebraic connectivity $\lambda_2$ is kept positive ($>0$) throughout the whole process, which encodes the fact that the network formed by the robots stayed connected. This implies that the CBF-derived connectivity maintenance constraint is playing its role well. More interestingly, under all three settings, the minimum achieved algebraic connectivity is slightly above 0. This is because the robots spread themselves out to approach the targets, such that the uncertainty in the targets' positions could be reduced. For the collision avoidance constraint (as shown in Fig.~\ref{fig:5i_4d_dist} to Fig.~\ref{fig:5i_6d_dist}), after only a few iterations, the minimum distance between robots is stabilized around $d_{\texttt{min}} = 3.0$, preventing the agents from getting too close or colliding into each other. Besides these intuitive illustrations of the system's performance, quantitative comparisons will be presented in the following section.

\subsection{Quantitative Results}

Discrete simulations for quantitative testing are run in a Python script. This allows for rapid and batched experiments, along with more effective parameter tuning. To evaluate the relative efficacy of the greedy algorithm (Algorithm~\ref{algo1}), we generated results comparing the performance of alternative control optimization methods. In these experiments, an equal number of robots and targets are randomly generated and placed in a fixed $10\times10\ m^2$ environment; the initial covariance for each target is set to 10$\textbf{I}_2$, the initial covariance for each robot is $\textbf{I}_2$, and the magnitude of the control inputs for the domain-discretized greedy and random algorithms is set to 1.5 $m$. All constraints, including connectivity maintenance, collision avoidance, and motion limiting, were omitted to the keep focus on the algorithms' characteristics. After measurements are computed and the first EKF update is performed, three methods of control input selection are compared over 30 trials. The first algorithm, named NLP, uses Pydrake~\cite{drake} to solve the non-linear optimization program with the objective of minimizing the total trace of covariance over an unlimited continuous action space. The second algorithm (Greedy) is the greedy algorithm described in Algorithm \ref{algo1}, and works by sequentially selecting actions from a predefined discrete control input set to minimize the trace of covariance. The third algorithm (Random) randomly selects a control input for each robot using the same predefined discrete control input set as the greedy algorithm. The resulting squared estimation error and trace of covariance as well as the computation time are compared across each method. 

It can be observed from Fig~\ref{fig:quant_cov_trace} that NLP consistently produces solutions with the lowest trace of covariance. Our greedy algorithm is able to perform close to NLP even with a limited action pool. Unsurprisingly, Random generates the worst control inputs, resulting in the largest uncertainty. 
Fig~\ref{fig:quant_comp_time} shows that the runtime of NLP increases exponentially with the number of robots, which makes NLP  infeasible to run in real-time for teams comprising a large number of robots. Meanwhile, Greedy runs in polynomial time and is several orders of magnitude faster than NLP.

To sum up, our greedy algorithm achieves a joint localization and tracking performance that is close in quality to NLP, is significantly better than Random, and runs much faster than NLP.   

\section{Conclusion}
\label{sec:conclusion}
This paper studies the problem of simultaneous self-localization of the robot team and tracking of multiple dynamic targets. We develop a two-staged approach that includes a greedy algorithm to minimize the uncertainty in the positions of robots and targets. CBFs are adapted to ensure connectivity of the robot communication network and to prevent inter-robot collisions. Gazebo simulations demonstrate that with the proposed approach, the robots closely follow the targets while maintaining connectivity and avoiding collisions in real-time. In addition, quantitative results show that compared with a non-linear optimization solver, our greedy algorithm achieves favorable accuracy at a significantly lower computational cost. 

This work opens up several future research directions. 
One direction is to explore the GPS-limited scenario, where some robots are denied access to GPS or the GPS signals they receive are noisy. In this setting, the robots have to rely on communications with peers for localization. The robustness of our system could be further verified if the robot team is able to adapt and recover when experiencing large localization errors induced by GPS denial. The second future avenue is to decentralize our framework~\cite{liu2022decentralized}, such that the system could deal with even larger team sizes through parallel computation and neighboring communications. Further, we consider extending the proposed framework to cope with adversarial targets that can plan strategically to undermine the task performance by compromising robots' sensors and/or communications~\cite{zhou2018resilient,zhou2022distributed,zhou2022robust}.

\bibliographystyle{IEEEtran}
\bibliography{IEEEabrv,refs}
\end{document}